\pdfoutput=1

\documentclass[11pt]{article}

\usepackage{ACL2023}
\usepackage{booktabs}
\usepackage{times}
\usepackage{latexsym}
\usepackage{graphicx} 
\usepackage{subcaption}
\usepackage[T1]{fontenc}

\usepackage[utf8]{inputenc}

\usepackage{microtype}

\usepackage{inconsolata}

\usepackage{hyperref}

\title{Tiny Titans: Can Smaller Large Language Models Punch Above Their Weight in the Real World for Meeting Summarization?
}

\author{Xue-Yong Fu$^*$, Md Tahmid Rahman Laskar$^*$\\ \textbf{Elena Khasanova, Cheng Chen, Shashi Bhushan TN} \\
          Dialpad Inc. \\
   Vancouver, BC, Canada \\ 
  \texttt{\{xue-yong,tahmid.rahman,elena.khasanova,cchen,sbhushan\}@dialpad.com}}

\begin{document}
\maketitle

\def\thefootnote{*}\footnotetext{\textbf{Equal Contributions. Sorted by the Last Name.}}\def\thefootnote{\arabic{footnote}}

\begin{abstract}
Large Language Models (LLMs) have demonstrated impressive capabilities to solve a wide range of tasks without being explicitly fine-tuned on task-specific datasets. However, deploying LLMs in the real world is not trivial, as it requires substantial computing resources. In this paper, we investigate whether smaller, compact LLMs\footnote{LLMs that have less than 2B parameters are referred to as Compact LLMs in this work.} are a good alternative to the comparatively Larger LLMs\footnote{LLMs that have at least 7B parameters are referred to as Larger LLMs in this work.} to address significant costs associated with utilizing LLMs in the real world. In this regard, we study the meeting summarization task in a real-world industrial environment and conduct extensive experiments by comparing the performance of fine-tuned compact LLMs (e.g., FLAN-T5, TinyLLaMA, LiteLLaMA) with zero-shot larger LLMs (e.g., LLaMA-2, GPT-3.5, PaLM-2). We observe that most smaller LLMs, even after fine-tuning, fail to outperform larger zero-shot LLMs in meeting summarization datasets. However, a notable exception is FLAN-T5 (780M parameters), which performs on par or even better than many zero-shot Larger LLMs (from 7B to above 70B parameters), while being significantly smaller. This makes compact LLMs like FLAN-T5 a suitable cost-efficient solution for real-world industrial deployment. 


%

\end{abstract}
\section{Introduction}

The instruction following capabilities have made it possible for LLMs to achieve impressive performance in zero-shot scenarios \cite{laskar2023systematicchatgpt,qin2023chatgpt,bang2023multitask}, which has also led to an increase in using LLMs to solve real-world problems. For instance, in tasks like meeting summarization, LLMs have been widely utilized in recent times due to their impressive zero-shot performance \cite{laskar-etal-2023-building}.  

However, despite the effectiveness of LLMs in summarization, deploying LLMs in the real world to generate meeting summaries would also lead to an increase in production costs. While fine-tuning smaller language models \cite{t5}, such as BART \cite{bart}, Pegasus \cite{pegasus}, etc. led to state-of-the-art results across various summarization datasets, these models require large annotated datasets for model training, which are often difficult to obtain in real-world business scenarios. Moreover, these smaller language models also do not have instruction-following capabilities \cite{zhang2023instruction}. Thus, they cannot be trained to properly follow specific instructions if there is a change in user requirements. 

GPT-4 \cite{openai2023gpt4} is an LLM proposed by OpenAI which is widely considered the best-performing LLM currently available \cite{chang2023surveyllm}. GPT-4 generated responses are also used to fine-tune various LLMs that are significantly smaller in size in comparison to it \cite{peng2023instructiongpt4}. Since using the GPT-4 API significantly increases the API usage cost \cite {laskar-etal-2023-building}, it is often not practical to use in real-world scenarios. 

In this regard, this paper studies whether compact/smaller LLMs can be fine-tuned in a way that can mimic the performance of GPT-4, while also significantly reducing the deployment cost of using LLMs in production for meeting summarization. More specifically, this paper aims to provide a comprehensive analysis of various smaller and larger LLMs, which includes larger LLMs like GPT-3.5 (i.e., ChatGPT\footnote{\url{https://openai.com/blog/chatgpt}}), PaLM-2 \cite{palm2}, LLaMA-2 \cite{touvron2023llama2}, as well as smaller LLMs like FLAN-T5 \cite{flant5}, TinyLLaMA \cite{zhang2024tinyllama}, etc.

Our experimental results show that most smaller LLMs, even after fine-tuning, fail to outperform larger zero-shot LLMs in meeting summarization datasets. However, a notable exception is a fine-tuned FLAN-T5-Large, which achieves performance on par with much larger LLMs (from 7B to more than 70B) used in zero-shot settings, while being significantly smaller. This makes smaller LLMs like FLAN-T5 a suitable cost-efficient LLM for real-world deployment. Our extensive experiments would give insights into the cost-effective utilization of LLMs for summarizing business meeting transcripts. 
Below, we summarize our major contributions in this paper:

\begin{enumerate}
    \item We conduct an extensive evaluation of smaller LLMs and compare their performance with larger LLMs in several meeting summarization datasets to address several limitations of using LLMs in the real world.
    \item To ensure a fair evaluation and address the possibility of data contamination, we utilize (i) one real-world Automatic Speech Recognition (ASR)-generated transcription data from real-world business meetings, and (ii) constructed a new version of the QMSUM \cite{zhong2021qmsum} dataset where the reference summaries are re-generated to keep them similar to our production requirement (this also helps us avoid the possibility of data contamination in LLM-generated responses). 

\item  Finally, we demonstrate the advantage of deploying smaller LLMs for real-world usage based on the analysis of performance (accuracy and latency), inference cost, and computational resource requirements. 
    
\end{enumerate}





\section{Related Work}

Fine-tuning language models \cite{bart,pegasus,t5} based on the transformer architecture \cite{DBLP:conf/nips/VaswaniSPUJGKP17} has led to state-of-the-art performance in various summarization datasets. Since these transformer-based language models require domain-specific fine-tuning for best results, obtaining in-domain labeled data in real-world settings is not trivial. However, the notable zero-shot abilities of LLMs in summarization \cite{laskar-etal-2023-building} have attracted attention for their potential use in practical summarization systems where in-domain labeled datasets are not available.

While zero-shot LLMs have demonstrated impressive performance in tasks that lack large annotated datasets \cite{laskar2023systematicchatgpt,qin2023chatgpt,bang2023multitask,jahan2023comprehensive}, utilizing LLMs in the real world also has several limitations. For instance, GPT-4 is currently regarded as the best-performing LLM in terms of various evaluation benchmarks. However, the API cost of using GPT-4 is significantly higher than of any other LLMs \cite{laskar-etal-2023-building}. While fine-tuned versions of less expensive closed-source LLMs could reach performance comparable to GPT-4, using fine-tuned versions of these LLMs for inference significantly increases the API cost\footnote{\url{https://openai.com/blog/gpt-3-5-turbo-fine-tuning-and-api-updates}}. Since these closed-source LLMs are only available through APIs, they pose potential privacy risks. 

To mitigate the above issues, various open-source LLMs have been proposed \cite{touvron2023llama,touvron2023llama2,jiang2023mistral,jiang2024mistral8x7b}. Some of the major advantages of using open-source LLMs are: (i) they are available for in-house deployment, (ii) they can be fine-tuned to achieve performance comparable to larger closed-source LLMs, and finally, (iii) the inference cost of using both zero-shot and fine-tuned versions are the same.  Thus, open-source LLMs could be a good alternative that addresses the limitations of closed-source LLMs. 

However, deployment of the open-source LLMs in a way that ensures customer satisfaction, i.e., high accuracy with low latency, would require expensive computing resources such as powerful GPUs with large memory capacity. In addition, fine-tuning larger LLMs also requires scarce and costly computing resources which may not be available in many industries. While various optimization techniques \cite{wan2023efficient} like low-bit quantization \cite{frantar2022gptq,dettmers2023qlora}, parameter-efficient fine-tuning \cite{hu2021lora}, etc. have been proposed recently to address the computational limitations, they often come with other issues, such as a drop in accuracy and an increase in latency.  

In this paper, we aim to address these issues by studying whether we can fine-tune smaller LLMs with instruction-following capabilities to mimic the performance of larger LLMs such as GPT-4 while ensuring low latency with minimized inference cost. 

\section{Our Methodology}


The objective of this research is to study whether instruction-following LLMs that are smaller in size can be effectively utilized in a real-world system for meeting summarization to ensure performance comparable to the state-of-the-art larger LLMs while minimizing the inference cost. For this purpose, we select LLMs that have fewer than 2B parameters as the targeted compact LLMs for performance analysis. Moreover, in real-world meeting summarization scenarios, users may have different requirements for the LLMs. For instance, some users may prioritize meeting summaries that are detailed and comprehensive, whereas others may prefer the meeting summaries to be short and concise. In such cases, the instruction following capability is important for the LLMs that would be deployed in production such that they can fulfill variations in user demands. Therefore, in this paper, we also evaluate the performance of LLMs based on a diverse set of instructions to generate \textit{(i) Long Summary}, \textit{(ii) Medium Length Summary}, and \textit{(iii) Short Summary}. We follow the work of \citet{laskar-etal-2023-building} for prompt construction and use their \textit{Summarization via Truncation} approach for each type of instruction. Below are the examples of the prompts for each case. 
 
\textbf{Long:} Generate a long and descriptive summary of the following conversation.

\textbf{Medium:} Generate a summary of the following conversation.

\textbf{Short:} Generate a very short and concise summary of the following conversation.

\section{Experiments}
In this section, we first present our models along with their implementation details. Next, we demonstrate the datasets we used for evaluation. Finally, we demonstrate our experimental findings. 
\subsection{Models} 
We use three compact LLMs that have less than 2B parameters and compare their performance with various larger LLMs (having at least 7B parameters). In the case of larger LLMs, some of them are closed-source (e.g., GPT-3.5, PaLM-2, etc.). When we use these closed-source LLMs, we use their respective APIs. All open-source LLMs are implemented using the HuggingFace library \cite{wolf2019huggingface}. 
Below, we describe the models that we study in this work. 

\subsubsection{Larger Zero-Shot LLMs}

\textbf{GPT-3.5:} It is an autoregressive LLM that leverages reinforcement learning from human feedback (RLHF) mechanism. 
It is the first backbone model behind ChatGPT and obtains impressive zero-shot performance across various tasks \cite{laskar2023systematicchatgpt}. 
We use the \textit{gpt-3.5-turbo-0613} model with the default parameters from OpenAI\footnote{\url{https://platform.openai.com/docs/models/}}. 

\textbf{PaLM-2:} PaLM-2 is an LLM \cite{palm2} developed by Google. It leverages the mixture of objectives technique \cite{palm2} and significantly outperforms the original PaLM \cite{chowdhery2022palm} model. We use the \textit{text-bison@002} model in \textit{Google's VertexAI\footnote{\url{https://cloud.google.com/vertex-ai/docs/generative-ai/model-reference/text}}} with the default parameters for PaLM-2. 

\textbf{LLaMA-2:} LLaMA-2 \cite{touvron2023llama2} is an open-source LLM developed by Meta. One major advantage of LLaMA-2 over the previously mentioned LLMs is that it is open-sourced and available for both research and commercial purposes. In this paper, we use the respective Chat versions of LLaMA-2 for all of its variations: 7B, 13B, and 70B from HuggingFace\footnote{\url{https://huggingface.co/meta-llama}} \cite{wolf2019huggingface} with the default parameters for inference. 

\textbf{Mixtral-8x-7B:} The Mixtral 8x7B \cite{jiang2024mistral8x7b} is a Sparse Mixture of Experts (SMoE) language model which has the same architecture as Mistral 7B \cite{jiang2023mistral}, but with the difference that each layer is composed of 8 feedforward blocks or experts. This architectural change has made it possible for each token to have access to 47B parameters while using only 13B active parameters during inference. We use it for zero-shot evaluation with its default parameters.

\subsubsection{Smaller Fine-Tuned LLMs}

{\textbf{FLAN-T5:} FLAN-T5 \cite{flant5} is an extension of the T5 \cite{t5} model. The T5 model treats each task as a sequence-to-sequence problem. While the architecture of FLAN-T5 is similar to the original T5 model, it leverages instruction fine-tuning instead of traditional fine-tuning. We use its 80M parameter small, 250M parameter base, and 780M parameter large versions from HuggingFace\footnote{\url{https://huggingface.co/docs/transformers/model_doc/flan-t5}} in our experiments with the learning rate set to $2e-5$. We run 10 epochs for FLAN-T5-Large and 20 epochs for Base and Small. 

\textbf{TinyLLama:} TinyLlama \cite{zhang2024tinyllama} is a compact 1.1B parameter language model that is built on the architecture of Llama-2 \cite{touvron2023llama2}. It is pre-trained on around 1 trillion tokens and leverages various techniques (e.g. FlashAttention \cite{dao2022flashattention,dao2023flashattention2}) to achieve better computational efficiency. We fine-tune it for 10 epochs with the learning rate of $1e-5$.

\textbf{LiteLLama:} LiteLLaMA\footnote{\url{https://huggingface.co/ahxt/LiteLlama-460M-1T}} is a 460M parameter LLM that is also developed based on the architecture of LLaMA-2 and trained over 1T tokens on part of the RedPajama\footnote{\url{https://huggingface.co/datasets/togethercomputer/RedPajama-Data-1T}} datasets. We fine-tune it for 20 epochs with the learning rate of $2e-5$.

\subsection{Datasets}

 \begin{table}[t!]
\setlength{\tabcolsep}{4pt} 
\centering
\tiny
\begin{tabular}{c|c|c}
\toprule 
 & \textbf{In-Domain Dataset} & \textbf{QMSUM-I Dataset}   \\   \cmidrule(r){2-3}
\textbf{Type} & \textbf{Train / Test} & \textbf{Train / Test}   \\ \midrule
\textbf{No. of Samples} & 1360 / 157  & 486 / 111 \\
\textbf{Avg. Words Per Transcript} & 600 / 620 & 8947 / 9461  \\
\textbf{Avg. Words Per Summary (Overall)} & 88 / 87 & 333 / 335 \\
\textbf{Avg. Words Per Summary (Long)} & 122 / 122 & 532 / 523  \\
\textbf{Avg. Words Per Summary (Medium)} & 76 / 77 & 303 / 307  \\
\textbf{Avg. Words Per Summary (Short)} & 60 / 61 & 170 / 173  \\

\bottomrule
\end{tabular}
\caption{{Evaluation Dataset Statistics. }}
\label{table:dataset_stat}
\end{table}


While one of our objectives is to build an LLM-based meeting summarization system that has instruction-following capabilities for real-world usage, there are no meeting summarization datasets currently available having different gold reference summaries corresponding to different instructions such as varying summary lengths or formats. Thus, to evaluate the performance of various LLMs, we constructed two datasets: (i) one dataset is based on our proprietary in-domain business conversation transcripts, and (ii) the other leverages an academic dataset.  
Below, we describe these datasets (also see Table \ref{table:dataset_stat} for more details).

\textbf{(i) In-Domain dataset:} This is a dataset collected from Dialpad\footnote{\url{https://dialpad.com/}} consisting of real-world business meetings. Since GPT-4 is found to be the best performing LLM in a wide range of tasks including meeting summarization \cite{laskar-etal-2023-building}, alongside its impressive capability as an annotator \cite{peng2023instructiongpt4}, we use it to generate the reference summaries depending on the \textit{Long}, \textit{Medium}, and \textit{Short} summary instructions. 

\textbf{(ii) The QMSUM\textsubscript{Filtered} dataset:} We use the filtered version \cite{laskar-etal-2023-building} of the QMSUM dataset \cite{zhong2021qmsum} to generate the meeting summaries. Since this dataset is not instruction-focused, we regenerate the reference summaries using GPT-4 with three types of instructions: \textit{Long}, \textit{Medium}, and \textit{Short}. Due to the variation in summary instructions, our instruction (I) focused version of QMSUM, denoted as QMSUM-I\footnote{To help facilitate future research, we have released the QMSUM-I dataset here: \url{https://github.com/talkiq/dialpad-ai-research/tree/main/tiny_titans}}, contains 3 times more instances than the original filtered version. 





\subsection{Results and Discussions}
For performance evaluation, we use ROUGE-1, 2, L (R-1, R-2, R-L) \cite{lin2004rouge}
as our evaluation metrics. 
Below, we present our findings. 


\begin{table*}
\centering
\small
\setlength{\tabcolsep}{4pt}
\begin{tabular}{l|ccc|ccc}
\toprule
\multicolumn{1}{c}{} & \multicolumn{3}{c}{\textbf{In-Domain Dataset}} & \multicolumn{3}{c}{\textbf{QMSUM-I Dataset}} \\
\cmidrule(r){2-4} \cmidrule(r){5-7}

\multicolumn{1}{c}{\textbf{Models}}
& \textbf{ROUGE-1} & \textbf{ROUGE-2} & \textbf{ROUGE-L} & \textbf{ROUGE-1} & \textbf{ROUGE-2} & \textbf{ROUGE-L} \\
\midrule

\textbf{GPT-3.5 (Zero-Shot)}& 49.55 & 24.61 & 36.12 & 38.63 & 13.17 & 21.83 \\
\textbf{PaLM-2-text-bison@002 (Zero-Shot)} & 48.32 & 23.61 & 35.59 & 39.76 & 12.29 & 21.14 \\
\textbf{LLaMA-2-7B (Zero-Shot)} & 47.37 & 20.41 & 30.93 & 35.67 & 10.14 & 18.57 \\
\textbf{LLaMA-2-13B (Zero-Shot)}& 47.07 & 21.37 & 31.58 & 32.93 & 9.69 & 18.06\\
\textbf{LLaMA-2-70B (Zero-Shot)}& 46.55 & 20.42 & 32.02 & 33.85 & 9.50 & 18.23 \\
\textbf{Mixtral-8x7B (Zero-Shot)}& 51.99 & 25.76 & 36.86 & \textbf{40.70} & \textbf{13.29 }&\textbf{ 21.96} \\
\midrule
\textbf{TinyLLaMA-1.1B (Fine-Tuned)}& 50.17 & 22.38 & 33.66 & 23.97 & 6.06 & 16.59 \\
\textbf{LiteLLaMA-460M (Fine-Tuned)}& 42.64 & 15.31 & 26.95 & 16.66 & 3.80 & 11.43 \\
\textbf{FLAN-T5-Small-80M (Fine-Tuned)}& 21.19 & 8.13 & 16.74 & 20.18 & 4.49 & 16.1 \\
\textbf{FLAN-T5-Base-250M (Fine-Tuned)}& 34.44 & 14.36 & 25.33 & 30.41 & 9.45 & 20.24\\
\textbf{FLAN-T5-Large-780M (Fine-Tuned)}&\textbf{ 56.14} &\textbf{ 29.42 }&\textbf{ 41.11 }& 34.03 & 11.31 & 20.92\\

\bottomrule

\end{tabular}
\caption{{Performance of LLMs on the In-Domain and QMSUM-I datasets.}}
\label{tab:results}
\end{table*}

\subsubsection{Performance on Benchmark Datasets} 
We show the results for both zero-shot LLMs and fine-tuned compact LLMs in Table \ref{tab:results}. 
Below, we summarize our observations:

    (i) We find that in both datasets, FLAN-T5-Large is the best-performing fine-tuned smaller LLM. Whereas Mixtral-8x7B is the best-performing zero-shot model among the larger LLMs. 

    (ii) We find that the ROUGE scores of all models are quite lower in the QMSUM-I dataset in comparison to our in-domain dataset. This is expected in the case of the fine-tuned models since the size of the training set in the QMSUM-I dataset is much smaller than our In-Domain dataset. 
  
    (iii) In zero-shot settings, we find that generally, the performance of GPT-3.5 and PaLM-2 are comparable to Mixtral. However, LLaMA-2-70B not only fails to outperform these larger models, it also fails to outperform its smaller variations in both datasets in several scenarios. 
    
    (iv) In the case of the fine-tuned LLMs, we find that except FLAN-T5-Large, the larger fine-tuned models perform much better than smaller ones. For instance, TinyLLaMA-1.1B outperforms LLMs that are smaller in size than it. However, it fails to outperform FLAN-T5-Large which has about 300M fewer parameters. 

    (v) In the case of FLAN-T5 models, we find that the FLAN-T5-Large-780M significantly outperforms its smaller variants: 80M and 250M. 

    (vi) While FLAN-T5-Large-780M performs the best in our In-Domain dataset, 
    it fails to outperform much larger zero-shot LLMs like GPT-3.5, PaLM-2, and Mixtral-8x7b (even though its performance is on par or better than LLaMA-2 models in various metrics) in the QMSUM-I dataset.

    (vii) As an explanation of the performance of FLAN-T5-Large, it should be noted that we use the default context length of 2048 tokens for FLAN-T5 since our objective is to build an efficient summarization model for deployment in a specific industry. Since the average transcript length in our in-domain dataset is about 600 words, most parts of the transcript in our in-domain dataset can be covered within the context window of FLAN-T5 models. However, this default context length is about 5 times lower than the average transcript length in QMSUM-I, which could be the possible reason behind its comparatively poorer performance on QMSUM-I. This indicates that in datasets that have smaller context lengths, FLAN-T5-Large could be very useful. Nonetheless, to further improve performance in datasets that have larger meeting lengths while ensuring limited computational usage, other approaches such as \textit{Summarization via Chapterization} \cite{laskar-etal-2023-building} can be investigated.

\subsubsection{Case Studies} 

In this section, we conduct some case studies to further investigate the performance of the best-performing smaller fine-tuned LLM: the FLAN-T5-Large model. Below, we demonstrate our findings:

\paragraph{(i) Case Study on Fine-Tuning Performance:} Since FLAN-T5 performed on par or even better than the zero-shot LLaMA-2 models in our previous experiment, in this section, we conduct a case study to compare its performance with the LLaMA-2-7B and LLaMA-2-13b models that are fine-tuned for 3 epochs with learning rate $2e-5$. We show our experimental results in Table \ref{table:case_study_fine_tuning_performance} and find that fine-tuning led to LLaMA-2 models (both 13B and 7B) outperforming FLAN-T5-Large in both datasets, with the improvement in QMSUM-I is by a large margin. The larger difference in performance in QMSUM can be attributed to the longer transcripts in QMSUM-I where the longer sequence length (context length of 4k tokens) in LLaMA-2 models could be more suitable than the context length of 2048 tokens in FLAN-T5-Large. Nonetheless, the improvements for fine-tuned LLaMA-2 models in our In-Domain dataset are quite narrow. 

 \begin{table}[t!]
\centering
\small
\setlength{\tabcolsep}{2pt}
\begin{tabular}{l|lll|lll}
\toprule 
  \multicolumn{1}{c}{} & \multicolumn{3}{c}{\textbf{In-Domain}} & \multicolumn{3}{c}{\textbf{QMSUM-I}} \\ 
\cmidrule(r){2-4} \cmidrule(r){5-7}
    \multicolumn{1}{c}{\textbf{Model}} & \textbf{R-1} &  \textbf{R-2} &   \textbf{R-L} &    \textbf{R-1} &  \textbf{R-2} &   \textbf{R-L}   \\  \midrule

 \textbf{FLAN-T5-Large} &{56.14} &{29.42}&{41.11}& 34.03 & 11.31 & 20.92\\
 \textbf{LLaMA-2-7B} & 57.09 &  30.42 & 41.68 & 42.77 & 13.93     &    22.16\\
  \textbf{LLaMA-2-13B} & 58.92 & 32.70 & 44.04 & 43.86 & 14.39     &  22.58\\

\bottomrule
\end{tabular}
\caption{{Results based on Fine-Tuning Smaller and Larger LLMs. }} 
\label{table:case_study_fine_tuning_performance}
\end{table}

\begin{figure*}[t!]
    \centering

        \includegraphics[height=4.32cm, width=14cm]{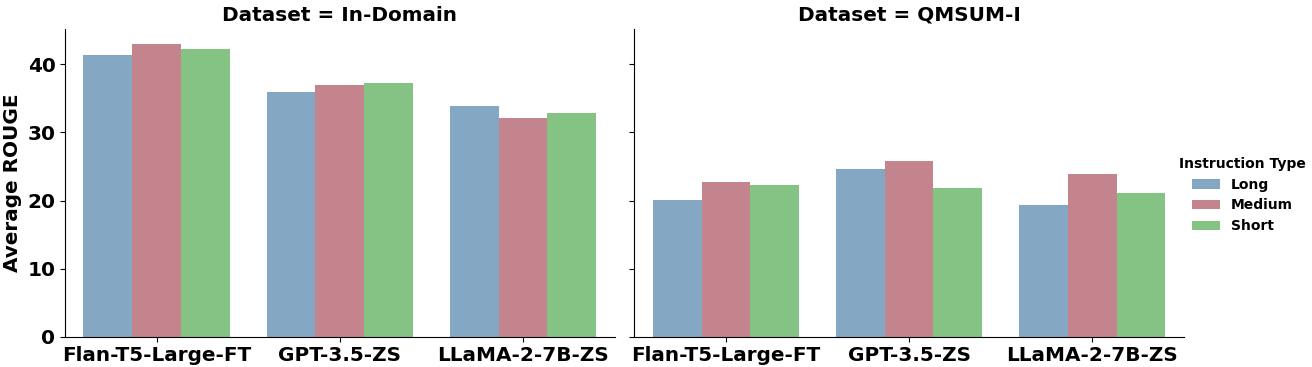}
    
    \caption{{Average ROUGE scores based on the instruction types for Fine-Tuned (FT) and Zero-Shot (ZS) LLMs.}}
    \label{fig:avg}
\end{figure*}

\paragraph{(ii) Case Study on Instruction Variations:} Here, we study the performance of some LLMs in terms of the variations in instructions. For the case study, we use the best-performing FLAN-T5-Large and compare it with two zero-shot larger LLMs, one API-based: GPT-3.5, and one open-source: LLaMA-2-7B\footnote{We select LLaMA-2-7B since it is the smallest one among all zero-shot LLMs, making it more suitable for deployment.}. We find that on our In-Domain dataset, FLAN-T5-Large performs better in Medium summaries, whereas GPT-3.5 and LLaMA-2-7B are better in Short and Long summaries, respectively. In QMSUM-I, we find that all LLMs perform the best in Medium summaries.

 \begin{table}[t!]
\centering
\small
\setlength{\tabcolsep}{5pt}
\begin{tabular}{l|lll|lll}
\toprule 
  \multicolumn{1}{c}{} & \multicolumn{3}{c}{\textbf{In-Domain}} & \multicolumn{3}{c}{\textbf{QMSUM-I}} \\ 
\cmidrule(r){2-4} \cmidrule(r){5-7}
    \multicolumn{1}{c}{\textbf{Model}} & \textbf{F} &  \textbf{C} &   \textbf{FC} &    \textbf{F} &  \textbf{C} &   \textbf{FC}   \\  \midrule

 \textbf{FLAN-T5-Large-FT} &{4.7} &{4.6}&{4.4}& 3.1 & 2.8 & 3.4\\
 \textbf{GPT-3.5-ZS} & 5.0 &  3.9 & 4.5 & 4.1 & 3.8     &    3.9\\
  \textbf{LLaMA-2-7B-ZS} & 4.8 & 3.5 & 3.3 & 3.8 & 3.4     &  3.9\\
\bottomrule
\end{tabular}
\caption{{Human Evaluation Results in terms of Fluency (F), Coherence (C), and Factual Consistency (FC). Here, `FT' denotes `Fine-Tuned', `ZS' denotes `Zero-Shot'.}} 
\label{table:human_eval}
\end{table}

\subsubsection{Human Evaluation Results} To provide more insights on LLM performance, we conduct a human evaluation to rate the LLM-generated summaries on a scale of 1 to 5 in terms of Fluency, Coherence, and Factual Consistency. We compare the best-performing smaller LLM: FLAN-T5-Large with two zero-shot baselines: GPT-3.5 and LLaMA-2-7B.  From the results in Table \ref{table:human_eval}, we find that similar to the performance in terms of ROUGE scores, all LLMs generally achieve better performance on our In-Domain dataset than the QMSUM-I dataset. We also find that on average, the performance of FLAN-T5-Large is better than GPT-3.5 and LLaMA-2-7B on our In-Domain dataset. Much longer meetings in the QMSUM-I dataset could be the reason behind FLAN-T5-Large performing poorly on this dataset. 

\section{Using LLMs in Real-World Systems}

To \textit{deploy} LLMs in the real world, we study the following aspects: \textit{cost/GPU} and \textit{inference speed}.

\textbf{Cost/GPU:} As of the time of writing this paper, the pricing\footnote{\url{https://openai.com/pricing}, last accessed: 01/25/2024.} in OpenAI for the GPT series models are as follows: the 4K context version of GPT-3.5 that we use costs $0.0015$\$ per 1K input tokens and $0.002$\$ per 1K output tokens.  Meanwhile, for PaLM-2, the pricing\footnote{\url{https://cloud.google.com/vertex-ai/pricing}, last accessed: 01/25/2024.} in Google Cloud is $0.00025$\$ per 1K characters and $0.0002$\$ per 1K output characters. Approximately, 1 token is considered as 4 characters. Thus, the cost for PaLM-2 is $0.0010$\$ per 1K input tokens and $0.0008$\$ per 1K output tokens, making it 
slightly cheaper than GPT-3.5. 
In terms of open-source LLMs (using 16-bit floating-point precision), we find that LLaMA-2-7B  
requires at least a machine with 1 NVIDIA L4 GPU (24GB VRAM), while the LLaMA-2-13B model requires 2 L4 GPUs (48GB VRAM). On the contrary, the 
FLAN-T5-Large-780M consumes about 6GB of VRAM. Thus, it can be run on much cheaper GPUs.   

\textbf{Inference Speed:} We also measure the inference speed of different LLMs in a machine having 1 L4 GPU. For this purpose, we use 100 transcripts consisting of real-world business conversations collected from \citet{laskar-etal-2023-building}. We find that on average, FLAN-T5-Large only takes 4.2 seconds per transcript, whereas LLaMA-2-7B takes 15 seconds per transcript \cite{laskar-etal-2023-building}.

\section{Conclusion} 
In this paper, our extensive study involving various LLMs led to several key insights on building an efficient meeting summarization system for real-world usage. While most larger LLMs usually outperform their smaller counterparts, we find that FLAN-T5-Large is an exception in this regard. On our In-Domain dataset, with only 780M parameters, FLAN-T5-Large not only outperforms larger zero-shot LLMs, but also it achieves comparable performance with larger fine-tuned LLMs. This makes FLAN-T5-Large more suitable for real-world usage, especially in scenarios where the meetings are not too long. 
Since the performance of FLAN-T5-Large is still quite below in comparison to other larger LLMs on QMSUM-I dataset that has longer meetings, future work should investigate the performance of FLAN-T5 by 
applying various chapterization techniques \cite{laskar-etal-2023-building}.  


\section*{Limitations}
One of the limitations of this work is that only three types of instructions were utilized. Thus, in the future, LLMs should be evaluated across more instructions. 

Another limitation of this work is that the GPT-4 generated summaries were utilized as reference summaries instead of human annotations. Nonetheless, one of the major focuses of this work is to ensure the efficient development of a real-world meeting summarization system. Since there is a lack of in-domain annotated datasets, we investigate the performance of different LLMs to mimic the performance of GPT-4 and so GPT-4 generated responses are utilized as the gold reference summaries. However, future work should evaluate the quality of GPT-4 generated summaries based on human evaluation. 

Another limitation that should be pointed out is that the performance of LLMs that were evaluated was based on truncating the transcript to the first $N$ tokens that can be covered by the maximum sequence length of the respective LLM. While this is done since the motivation of this work was to build an efficient summarization system that may reduce the production cost in a real-world industrial environment (note that our in-domain dataset also has shorter meetings), future work should investigate the performance of smaller LLMs by applying various chapterization techniques.

Finally, studying the effects of the size of the datasets used for fine-tuning smaller LLMs were left out of the scope of this work and will need to be considered in future research.


\section*{Ethics Statement}

\paragraph{License:} We maintained the licensing requirements accordingly while using different tools from the providers (e.g., OpenAI, Google, Meta, Mistral, HuggingFace). 

\paragraph{Privacy:} To protect user privacy, sensitive
data such as personally identifiable information (e.g., credit card number, phone number, person names)
were removed while constructing the In-Domain datasets. 

\paragraph{Intended Use:}  Note that our model is intended to provide business organizations with a quick overview of the meetings. While poor summarization quality may lead to a bad user experience, it should not lead to any ethical concern since the summary is required to be generated based on only the given transcript. Meanwhile, the LLM that would be used in production for summarization will only do inference but will not be re-trained on live meeting transcripts. Only the users of a particular meeting will have access to the summary. Thus, information from any other meetings will not be revealed to the users. 

\paragraph{Human Evaluation:} Additional compensations were not required for the human evaluation since it was conducted by in-house full-time employees having expertise in computational linguistics.

\bibliography{anthology,custom}
\bibliographystyle{acl_natbib}

\end{document}